%% file: anonymous-submission-latex-2025.tex
\documentclass[letterpaper]{article} % DO NOT CHANGE THIS
\usepackage[submission]{aaai25}  % DO NOT CHANGE THIS
\usepackage{times}  % DO NOT CHANGE THIS
\usepackage{helvet}  % DO NOT CHANGE THIS
\usepackage{courier}  % DO NOT CHANGE THIS
\usepackage[hyphens]{url}  % DO NOT CHANGE THIS
\usepackage{graphicx} % DO NOT CHANGE THIS
\def\UrlFont{\rm}  % DO NOT CHANGE THIS
\usepackage{natbib}  % DO NOT CHANGE THIS AND DO NOT ADD ANY OPTIONS TO IT
\usepackage{caption} % DO NOT CHANGE THIS AND DO NOT ADD ANY OPTIONS TO IT
\usepackage{algorithm}
\usepackage{algorithmic}
\usepackage{amsmath} 
\usepackage{amssymb}
\usepackage{adjustbox}
\usepackage{subcaption}
\usepackage{multirow}
\usepackage{booktabs}
\usepackage[table]{xcolor}

\usepackage{xcolor}

\usepackage{newfloat}
\usepackage{listings}
\DeclareCaptionStyle{ruled}{labelfont=normalfont,labelsep=colon,strut=off} % DO NOT CHANGE THIS
\floatstyle{ruled}
\newfloat{listing}{tb}{lst}{}
\floatname{listing}{Listing}
\title{scFusionTTT: Single-cell transcriptomics and proteomics fusion with Test-Time Training layers}
\author{
    Dian Meng\textsuperscript{\rm 1}\equalcontrib,
    Bohao Xing\textsuperscript{\rm 1}\equalcontrib,
    Xinlei Huang\textsuperscript{\rm 1},
    Yanran Liu\textsuperscript{\rm 1},
    Yijun Zhou\textsuperscript{\rm 1},
    Yongjun xiao\textsuperscript{\rm 1},
    Zitong Yu\textsuperscript{\rm 1},
    Xubin Zheng\textsuperscript{\rm 1}\thanks{Corresponding author.}
}
\affiliations{
    \textsuperscript{\rm 1}School of Information Science and Technology, Great Bay University\\
    xbzheng@gbu.edu.cn
}
\usepackage{bibentry}
\begin{document}

\maketitle

\begin{abstract}
Single-cell multi-omics (scMulti-omics) refers to the paired multimodal data, such as Cellular Indexing of Transcriptomes and Epitopes by Sequencing (CITE-seq), where the regulation of each cell was measured from different modalities, i.e. genes and proteins. scMulti-omics can reveal heterogeneity inside tumors and understand the distinct genetic properties of diverse cell types, which is crucial to targeted therapy. Currently, deep learning methods based on attention structures in the bioinformatics area face two challenges. The first challenge is the vast number of genes in a single cell. Traditional attention-based modules struggled to effectively leverage all gene information due to their limited capacity for long-context learning and high-complexity computing. The second challenge is that genes in the human genome are ordered and influence each other's expression. Most of the methods ignored this sequential information. The recently introduced Test-Time Training (TTT) layer is a novel sequence modeling approach, particularly suitable for handling long contexts like genomics data because TTT layer is a linear complexity sequence modeling structure and is better suited to data with sequential relationships. In this paper, we propose scFusionTTT, a novel method for \textbf{S}ingle-\textbf{C}ell multimodal omics \textbf{Fusion} with \textbf{TTT}-based masked autoencoder. Of note, we combine the order information of genes and proteins in the human genome with the TTT layer, fuse multimodal omics, and enhance unimodal omics analysis. Finally, the model employs a three-stage training strategy, which yielded the best performance across most metrics in four multimodal omics datasets and four unimodal omics datasets, demonstrating the superior performance of our model. The dataset and code will be available on \url{https://github.com/DM0815/scFusionTTT}. 
\end{abstract}
\section{Introduction}
Cellular Indexing of Transcriptomes and Epitopes by Sequencing (CITE-seq) \cite{stoeckius2017simultaneous} is a technology that can measure transcriptomics and proteomics in individual cells simultaneously. 
The joint analysis of transcriptomics and proteomics can strengthen critical genetic information from multiple omics and unravel cellular processes.

\begin{figure}[t]
  \centering
  \includegraphics[width=0.9\linewidth]{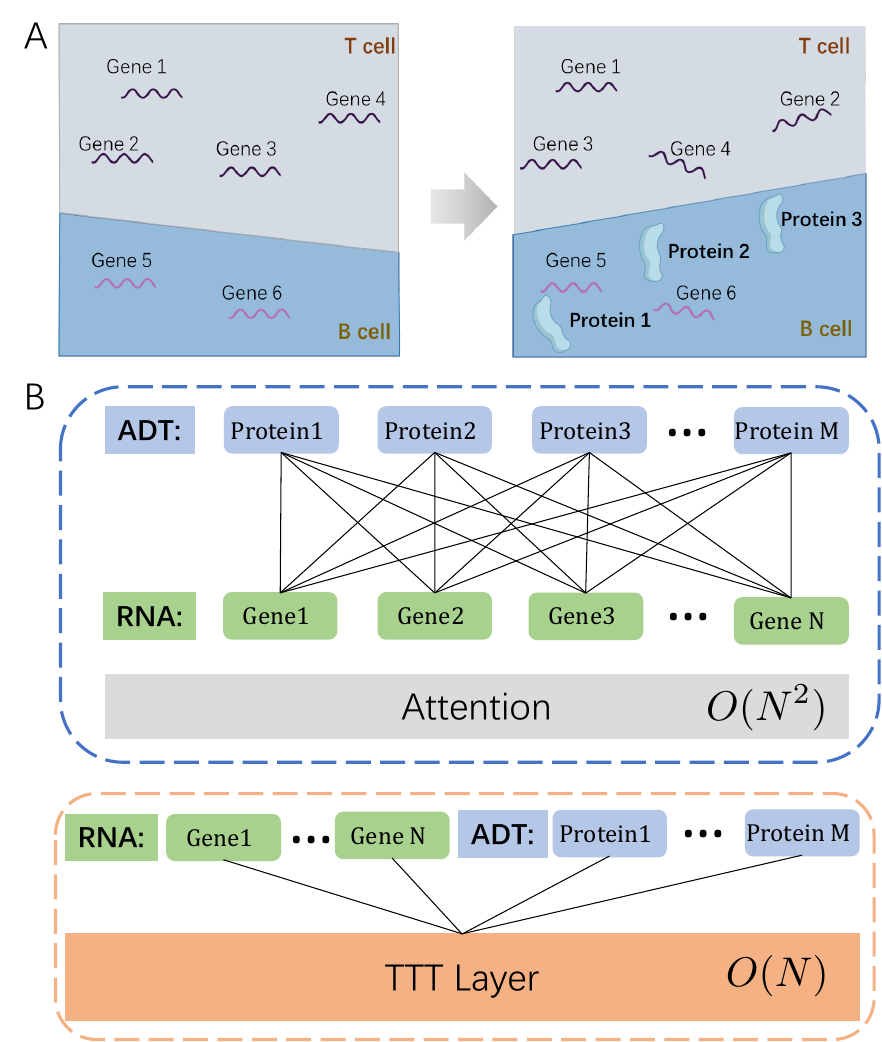}
  \caption{(A)Adding more modalities can aid cellular analysis by providing separate information from various omics. (B) Comparison of complexity and computation between attention and TTTlayer when conducting single-cell multi-omics fusion.}
  \vspace{-1em}
  \label{Fig1}
\end{figure}

Although fusing scMulti-omics data can help researchers explore complex biological information, the intrinsic properties of single-cell data, such as high sparsity, noise, and dimensionality mismatch, pose significant computational and analytical hurdles. Therefore, deep learning methods and variational inference methods have emerged as the mainstream techniques for scMulti-omics data analysis to solve above problems. Taking CITE-seq as an example, some series of algorithms have recently been proposed for it. scCTCLust \cite{yuan2022clustering} fuses transcriptomics and proteomics data utilizing Variational Autoencoders (VAE) \cite{doersch2016tutorial} and canonical correlation analysis. TotalVI \cite{steier2022joint} based on variational inferencing takes batch effects and protein background into consideration. Recently, a probabilistic tensor decomposition method \cite{wang2023probabilistic} is proposed to fuse scMulti-omics.

However, the models described above ignore the distinct features inherent in each omic, especially in Bayesian-based methods. The various features of different omics can provide additional information for cellular investigation. For instance, while the expression abundance of genes may provide several possibilities for cell classification, the unique information received by proteomics may provide heightened confidence, thus boosting the efficacy of cellular identification as shown in Figure \ref{Fig1}A. 

Nevertheless, the typical attention-based module has significant token number limitations and cannot effectively exploit the anterior and posterior gene information, because of the enormous number of genes in each sample. A recent proposed sequence modeling, Test-Time Training (TTT) layer \cite{sun2024learning}, used self-supervised learning as the hidden state updating rules, replacing traditional RNN hidden states updating, similarity calculation of attentional mechanisms, and achieving context compression through gradient descent of input tokens. Unlike the previous attention-based models, the TTT layer can unlock linear complexity structures via expressive memory, resulting in greater performance in lengthy contexts, which is ideal for genomic data as shown in Figure \ref{Fig1}B.

Especially, genes are sequentially arranged on human chromosomes, and the order of genes can affect the expression of nearby genes \cite{dandekar1998conservation, hurst2004evolutionary, wittkopp2012cis}. 
In addition, different proteins are transcribed and translated from different genes, genes can directly regulate each other, resulting in protein expression affecting one another \cite{maston2006transcriptional}. 
Most models often overlook this important aspect.

\begin{figure*}
  \centering
  \vspace{-0.5em}
  \includegraphics[width=0.95\textwidth]{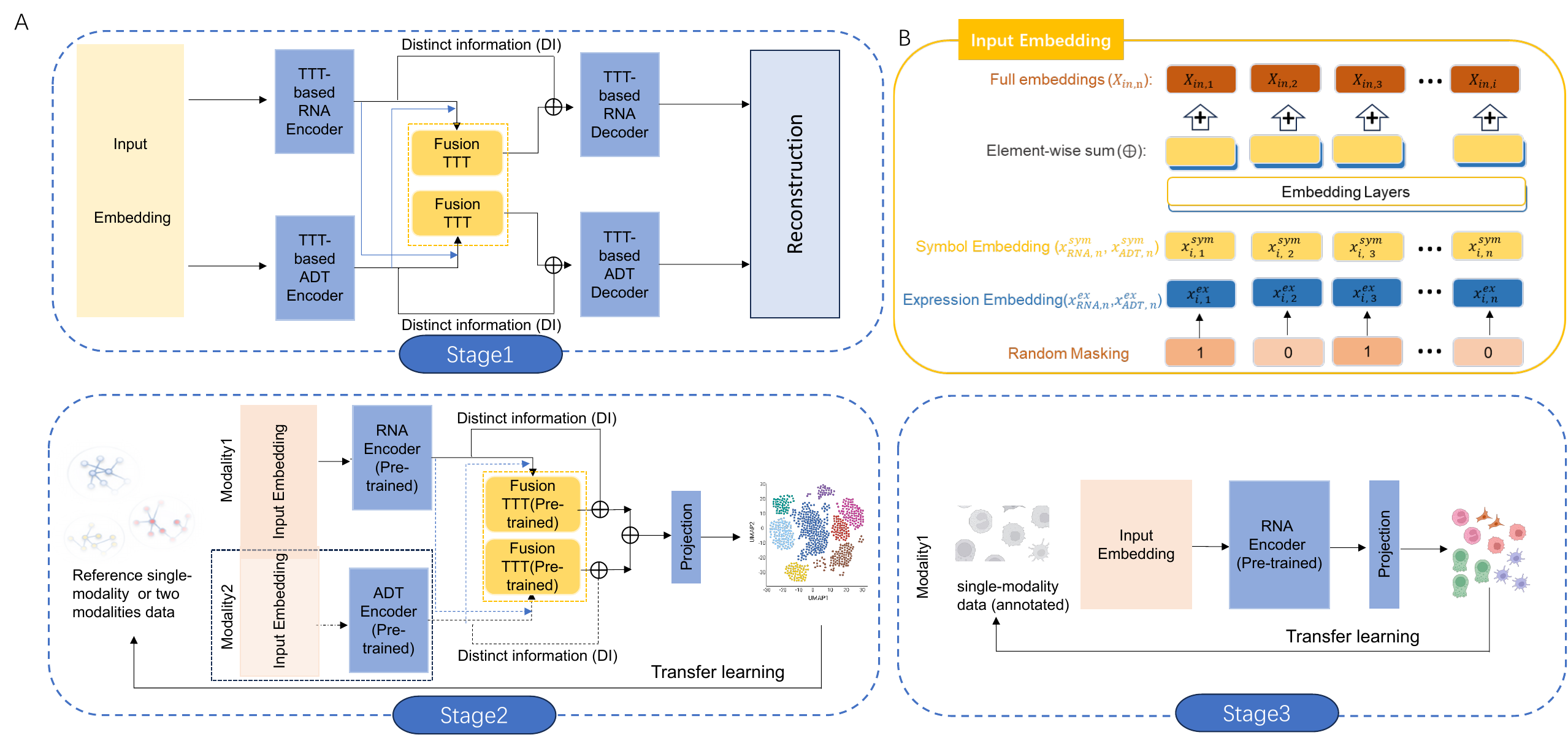}
  \vspace{-1em}
  \caption{ Overview of scFusionTTT. (A) The overall model consists of three stages to learn the latent representations of cells of multimodal omics and unimodal omics. (B) The input of scFusionTTT consists of expression embedding and symbol embedding.}
  \vspace{-1.5em}
  \label{workflow} 
\end{figure*}

Motivated by the above observations, we propose scFusionTTT, a scMulti-omics fusion method via masked autoencoder with TTT layers, as described in Figure \ref{workflow}A. In order to make the TTT layer fit better with the omics data, we added gene and protein order information in the human genome to the model.  A masked TTT-based autoencoder was applied for model pretraining and was used to transfer knowledge to enhance single-cell transcriptomics utilizing transfer learning.

The main contributions of our work are summarized below:
\begin{itemize}
\item We proposed a scMulti-omics including transcriptomics and proteomics fusion model called scFusionTTT, combining the TTT-based masked autoencoder and took the gene and protein order information in the human genome into consideration.

\item  We transferred multi-omics knowledge learnt from scFusionTTT to unimodal omics analysis.

\item  To the best of our knowledge, this is the first model proposing Fusion methods with TTT layers and applied to single-cell multi-omics analysis area.

\item  We compared our model with competitive state-of-the-art method on 4 CITE-seq datasets and 4 scRNA-seq datasets. The results demonstrated that scFusionTTT outperforms all the other baseline methods.
\end{itemize}
\section{Related Work}
\subsection{Single-cell Multi-omics Fusion}
Multi-omics data provide researchers with a comprehensive understanding of biological systems from various perspectives, as these different omics have complementary roles and collaborate to perform specific biological functions. However, multi-omics data are complex, high-dimensional, and heterogeneous, making it challenging to extract valuable insights from them \cite{boehm2022harnessing, miao2021multi}. Various approaches have been developed to address this challenge, such as multi-kernel learning, Bayesian consensus clustering, machine learning-based dimensionality reduction, similar network fusion, and deep learning methods.

% In this section, we present recent and mainstream work on the single-cell multi-omics dataset. Currently, most methods for the integration analysis of transcriptomics and proteomics data are based on probabilistic graphical models, which can be further divided into two subcategories. One subcategory includes models based on Variational Autoencoders (VAE) \cite{} and their variants, such as scCTCLust and scMM. scCTCLust [6] integrates transcriptomic and proteomic data from a single cell utilizing VAE and canonical correlation analysis. ScMM \cite{} is specifically designed for the analysis of CITE-seq data, emphasizing joint representation and predictions across different modalities. TotalVI \cite{} presents an end-to-end framework for the joint analysis of CITE-seq data. It probabilistically characterizes the data by integrating both biological and technological factors, such as protein background and batch effects. The other category comprises models that utilize traditional Bayesian methods with Gibbs sampling. For instance, jointDIMMSC \cite{} and BREMSC \cite{} are Bayesian Random Effects Mixture Models designed for joint clustering of CITE-Seq data. These models employ Gibbs sampling to sample from the posterior distribution given the observed data. Recently, the SCOIT model \cite{} was introduced as a probabilistic tensor decomposition framework, specifically designed to extract embeddings from paired single-cell multi-omic data, including CITE-seq data.

\subsection{RNNs and Test-Time Training}
% RNN is a type of neural network architecture designed to handle sequential data. Unlike traditional feedforward neural networks, RNNs have cyclic connections that allow information to flow between time steps \cite{}. LSTM \cite{} GRU \cite{}.  

%  However, TTT \cite{}

RNN\cite{sherstinsky2020fundamentals} is a type of neural network architecture designed to handle sequential data. However, traditional RNNs like LSTM\cite{sherstinsky2020fundamentals} struggle with long-range dependencies, leading to issues like vanishing gradients. Recent models, such as Mamba\cite{gu2023mamba} and RWKV\cite{peng2023rwkv}, have improved RNNs by enhancing hidden state representations and using chunk-wise parallelism. Despite these advances, RNNs still face challenges when dealing with very long contexts, often showing diminished performance as context length increases. 

Test-Time Training \cite{sun2024learning} introduces a dynamic approach where the model’s hidden state is updated during inference through self-supervised learning. This allows the model to adapt to new data on the fly, making it better suited for handling unseen or out-of-distribution inputs. Based on this, TTT layers enhance RNNs in long-context processing by continuously updating the hidden state with incoming data. This approach has demonstrated superior performance in long-context tasks compared to both traditional RNNs \cite{salehinejad2017recent} and Transformers \cite{vaswani2017attention}.

\section{Method}
\subsubsection{Problem Definition}
Denote the CITE-seq samples set $S = \left\{ X_{\text{RNA}}^k, X_{\text{ADT}}^k \right\}, \ k \in [1,n]$, where k is the cell number. In the matrix form, $X_{\text{RNA}}^k\in \mathbb{R}^{n \times o}$ 
and $X_{\text{ADT}}^k\in \mathbb{R}^{n \times p}$ represent tanscriptomics and proteomics data, respectively. 
In transcriptomics, $o$ is the number of genes, and in proteomics, $p$ is determined by the number of proteins contained in each dataset. The goal of scFusionTTT is to learn a unified representation $Z^{n \times d}$ of transcriptomics and proteomics for each sample in $S$ integrating all the omics. To learn the embedding, we designed a Test-Time Training (TTT)-based RNA and (Antibody-Derived Tag) ADT masked autoencoder, RNA and ADT decoder, and two FusionTTT modules.

\subsubsection{TTT Layer with CITE-seq}

\begin{figure}
  \centering
  \includegraphics[width=\linewidth]{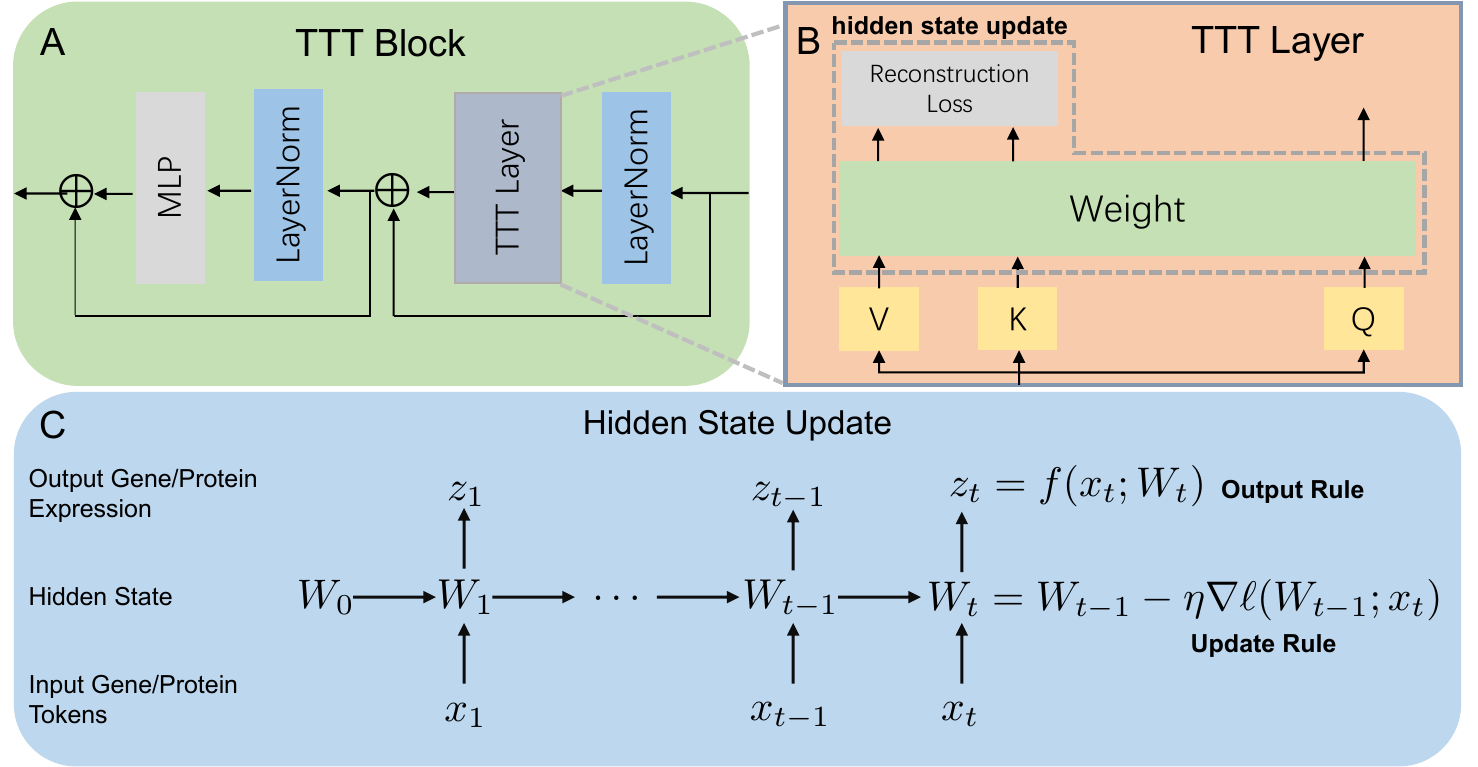}
  \vspace{-2em}
  \caption{(A) TTT Block. (B) TTT Layer with CITE-seq. (C) Pipeline of hidden state update.}
  \vspace{-1.5em}
  \label{TTTLinear}
\end{figure}

We leverage the self-supervised reconstruction process of TTT Layer to compress genetic sequence information into the model’s weights, as shown in Figure~\ref{TTTLinear}B. We introduce this process by taking RNA sequences $X_{\text{RNA}} = [x_t], t \in \{1, \dots, n\}$ as input. The process begins with an initial hidden state \( W_0 \), which is then iteratively refined using self-supervised learning. For each input gene token \( x_t \) in a single cell, the model computes an output:
\begin{equation}
 z_t = f(x_t; W_t)
\end{equation}
where \( W_t \) represents the current hidden state, and \( z_t \) is the updated gene or protein expression. This hidden state \( W_t \) is updated by utilizing a gradient descent step on a self-supervised loss \( \ell \) that aims to minimize the difference between the model's expression prediction and the actual expression. The update rule as shown in Figure~\ref{TTTLinear}C, also can be expressed as:
\begin{equation}
 W_t = W_{t-1} - \eta \nabla \ell(W_{t-1}; x_t)
\end{equation} 
where \( \eta \) is the learning rate. TTT layers further utilize a multi-view reconstruction \cite{chen2020improved} to compress more important information. 
The new self-supervised loss can be described as:
\begin{equation}
\ell(W; x_t) = \left\| f(\theta_K x_t; W) - \theta_V x_t \right\|^2
\end{equation} 
where \(\theta_K\) and \(\theta_V\) are the projections for the training view and label view, respectively. The corresponding output rule is as follows:
\begin{equation}
z_t = f\left(\theta_Q x_t; W_t\right)
\end{equation} 
where \(\theta_Q\) is the projection for the test view. The linear complexity and sequence modeling advantages of the TTT layer enable us to better model gene and protein sequences.
% Of note is that we refer to this entire self-supervised learning process as Test-Time Training (TTTLayer) as shown in Figure \ref{workflow}D.

% Through this iterative process, the model effectively "learns" from the incoming data at test time, allowing it to adapt to the specific characteristics of the test sequence, thereby improving its performance on tasks that involve long contexts and complex dependencies.

\subsubsection{RNA and ADT Masked Autoencoder}
The Test-Time Training Block (TTTBlock), composed of the TTTlayer, is the backbone of our RNA and ADT encoder, RNA and ADT decoder, and FusionTTT module in our model. Each encoder consists of two TTTBlocks, an MLP layer, and two Root Mean Square Layer Normalization (RMSNorm) \cite{zhang2019root} layers as depicted in Figure \ref{TTTLinear}A. The input $X_{in}$ to  the ADT and RNA encoder consists of gene expression embedding $X_{RNA}^{ex}$, gene symbol embedding $X_{RNA}^{sym}$, protein expression embedding $X_{ADT}^{ex}$, and protein symbol embedding $X_{ADT}^{sym}$ as shown in Figure \ref{workflow}B, which can be formulated as follows:
 \begin{equation}
 X_{in} = (X_{RNA}^{ex} + X_{RNA}^{sym} , X_{ADT}^{ex} + X_{ADT}^{sym}).
 \end{equation}

We initially randomly masked gene and protein expression values. Subsequently, the unmasked transcriptomics and proteomics data, that is, the unmasked features $X_{in}^{un}$, were encoded separately using a TTTBlock based on a self-supervised learning mechanism. The encoding process can be described as follows:
% \begin{equation}
% E^{out} = \sigma(\gamma(X_{in})) = (\theta_V*X_s)(\theta_K*X_s)^T(\theta_Q*X_t)
% \end{equation}
\begin{equation}
\begin{aligned}
X'_{l} &= \operatorname{TTT Layer}(\operatorname{\psi}(X_{l-1})) + X_{l-1}, \\
X_{l} &= \operatorname{MLP}(\operatorname{\psi}(X'_{l})) + X'_{l}.
\end{aligned}
\end{equation}
Where $\psi$ is the RMSNorm function, $X_0=X_{in}$ as the initial input of our model, and $X_l$ represents the outputs of the $l$-th TTTBlock. The overall block structure is similar to a standard self-attention block, with the key difference being that the self-attention is replaced by a TTTlayer, which leverages self-supervised learning to update hidden states. Therefore, the process of Encoder can be described as :
\begin{equation}
% D_{RNA}^{out}  = \gamma(D_{RNA}^{in})  
E_{RNA}^{out}  = \operatorname{ENC}_{RNA}(E_{RNA}^{in}),
\end{equation}
\begin{equation}
% D_{ADT}^{in}  = \gamma(D_{ADT}^{in})  
E_{ADT}^{in}  = \operatorname{ENC}_{ADT}(E_{ADT}^{in}).
\end{equation}
where $\operatorname{ENC}$ is the encoder, $E_{RNA}^{in}$, $E_{ADT}^{in}$, $E_{RNA}^{out}$, and $E_{ADT}^{in}$ represent the input to the RNA encoder, the input to the ADT encoder, the output of the RNA encoder, and the output of the ADT encoder, respectively.
% where $\sigma$ refers to the RMSNorm layer, $\gamma$ is a function that represents the TTT layer, $\theta_Q$, $\theta_K$, and $\theta_V$ are the paratermers of TTT layer, and $X_s$ $X_t$
%X_t X_s

% Usually, in the attention mechanism, the process can be calculated as follows:
% \begin{equation}
% E_t = (V_t)\operatorname{Softmax}((K_t)^T(q_t))= \sum_{s=1}^{t} v_s k_s^T q_t
% \end{equation}

%% 要把 隐藏层，updated state规则写下吗
\subsubsection{Multi-modalities Information Fusion}
To generate a unified representation of different modal omics, we introduce a fusion module to TTT. We design a FusionTTT module to integrate the outputs from each sample at the encoder stage and feed them into the fusion module. Given the construction of the TTT module, it is evident that the final token contains the most information. Therefore, we concate the outputs of the two encoders and pass them through the FusionTTT module, resulting in the outputs of the respective modalities.
\begin{equation}
\operatorname{FusionTTT}(E_{M_1}^{out}, E_{M_2}^{out}) = \operatorname{TTT Block}([E_{M_1}^{out}, E_{M_2}^{out}]),
\end{equation}
where $[...\, , \,...]$ is the concatenation operation, $E_{M_1}^{out}$ and $E_{M_2}^{out}$ are the outputs of the encoder in their respective modal, which shows that the modalities can be easily extended to more than three modalies.  In this experiment, \( E_{M_1}^{out} = E_{RNA}^{out} \) and \( E_{M_2}^{out} = E_{ADT}^{out} \). If this fusionTTT module is targeting a modality, then the output of the encoder for that modality is placed later.
Due to the sequence modeling capability of TTT, we think that this approach can transfer information from \( E_{M_1}^{out} \) to \( E_{M_2}^{out} \). Therefore, $E_{RNA}^{out}$ and $E_{ADT}^{out}$ will be contacted to one sequence, and fed into the two FusionTTT modules, respectively. The process is as follows:
% \begin{equation}
% FT_{RNA}^{in} = E_{RNA}^{out} + \lambda(E_{ADT}^{out}, E_{RNA}^{out})
% \end{equation}
% \begin{equation}
% FT_{ADT}^{in}  = E_{ADT}^{out} + \lambda(E_{RNA}^{out}, E_{ADT}^{out})
% \end{equation}
% where $\lambda(...)$ is function that cat the elements inside (...).
\begin{equation}
FT_{RNA}^{in} = E_{RNA}^{out} + \lambda \cdot \operatorname{FusionTTT}{(E_{ADT}^{out}, E_{RNA}^{out})},
\end{equation}
\begin{equation}
FT_{ADT}^{in}  = E_{ADT}^{out} + \lambda \cdot \operatorname{FusionTTT}{(E_{RNA}^{out}, E_{ADT}^{out})},
\end{equation}
where $\lambda$ is a learnable hyperparameter.
% %%%再过下TTT module,用一个函数表示？
% \begin{equation}
% FT_{RNA}^{out} = \gamma(FT_{RNA}^{in})
% \end{equation}
% \begin{equation}
% FT_{ADT}^{out}  = \gamma(FT_{ADT}^{in})
% \end{equation}

\subsubsection{RNA and ADT Decoder}
After each modality has learned the information of the other modalities, we go through the respective modality decoder to reconstruct the initial matrices of the respective modalities. At first, 
the input of RNA and ADT decoder that can be described as:
\begin{equation}
D_{RNA}^{in} = E_{RNA}^{out} + FT_{RNA}^{out},
\end{equation}
\begin{equation}
D_{ADT}^{in}  = E_{ADT}^{out} + FT_{ADT}^{out}.
\end{equation}
Next, RNA and ADT decoders will be used to reconstruct the transcriptomics and proteomics, respectively.
\begin{equation}
% D_{RNA}^{out}  = \gamma(D_{RNA}^{in})  
D_{RNA}^{out}  = \operatorname{DEC}_{RNA}(D_{RNA}^{in}),
\end{equation}
\begin{equation}
% D_{ADT}^{in}  = \gamma(D_{ADT}^{in})  
D_{ADT}^{in}  = \operatorname{DEC}_{ADT}(D_{ADT}^{in}).
\end{equation}
Where $\operatorname{DEC}$ is the decoder, $D_{RNA}^{out}$, and $D_{ADT}^{in}$ are the outputs of RNA and ADT decoder, respectively.
Of note, we removed the decoder module and just used the results of FusionTTT module as the final representation in stage 2 and stage 3. 

\input{citeseq_tab}

\subsubsection{Loss Function}
The overall training process of scFusionTTT contains three stages,  namely, multi-omics pre-training stage, multi-omics fine-tuning stage, and the unimodal omics predicting stage as shown in Figure \ref{workflow}A. In the multi-omics pre-training stage, the loss function for training is defined as :
\begin{equation}
\begin{split}
     \mathcal{L}_1 = \alpha \left( \frac{1}{n} \sum_{i=1}^{n} (X_{RNA,i}^{ex} - D_{RNA,i}^{out})^2 \right) + \\
     \beta \left( \frac{1}{n} \sum_{i=1}^{n} (X_{ADT,i}^{ex} - D_{ADT,i}^{out})^2 \right),
\end{split}
\end{equation}
where $\alpha$ and $\beta$ are the hyper-parameters used to determine the loss weights for transcriptomics and proteomics independently, and $n$ is the cell number in experiments. $X_{RNA,i}^{ex}$, $X_{ADT,i}^{ex}$, $D_{RNA,i}^{out}$, and $D_{ADT,i}^{out}$ are the elements of $X_{RNA}^{ex}$, $X_{ADT}^{ex}$, $D_{RNA}^{out}$, and $D_{ADT}^{out}$, which represents gene true expression values, gene predicted values, protein true expression values and protein predicted values, respectively. 

After scFusionTTT learned the gene and protein expression information, the model will learn the cell type information in the multi-omics fine-tuning stage utilizing cross-entropy loss, which can be described as follows:
\begin{equation}
     \mathcal{L}_2 =  \frac{1}{n} \sum_{i=1}^{n} \left[ y_i \log(\hat{y}_i) + (1 - y_i) \log(1 - \hat{y}_i) \right]^{}.
\end{equation}
where $n$ is the cell number of CITE-seq data in the training process, $y_i$ and $\hat{y}_i$ are the true cell type labels and predicted cell type labels, respectively.

Finally, we transfer multi-omic knowledge to transcriptomics in the unimodal omics predicting stage. The loss is calculated as follows:
\begin{equation}
     \mathcal{L}_3 = \frac{1}{m} \sum_{i=1}^{m} \left[ y_i \log(\hat{y}_j) + (1 - y_j) \log(1 - \hat{y}_j) \right]^{}.
\end{equation}
Where $m$ is the cell number of transcriptomics data in the training process, $y_j$ and $\hat{y}_j$ are the true cell type labels and predicted cell type labels, respectively.

\section{Experiments} 
\subsubsection{Data Sources and Prepocessing}
We utilized four CITE-seq datasets, and four RNA-seq datasets at experiments, and detailed information regarding these datasets was provided in Supplementary materials.

We applied distinct normalization strategies tailored to each data type: reads per kilobase per million (RPKM) \cite{conesa2016survey} normalization for transcriptomics data, and Centered Log Ratio (CLR) \cite{aitchison1982statistical} normalization for the proteomic data to mitigate compositional effects.
\begin{equation}
\text{CLR}(x_i) = \log \left( \frac{x_i}{\sqrt[n]{\prod_{j=1}^{n} x_j}} \right),
\end{equation}
where $x_i$ represents the $i$-th protein expression value in the cell, with $n$ denoting the total number of proteins in a cell, and $j$ iterating over all proteins.
After normalization, we selected 4000 high variable genes along with all proteins for model input to capture the most informative features. All preprocessing procedures were executed using Scanpy’s integrated functions.

%See Supplementary Methods S1.1 and S1.2 for procedures and parameterization of multi-omics and unimodal omics, respectively.
\subsubsection{Training Strategy} 
The initial step in scFusionTTT involves reconstructing RNA and protein matrices for transcriptomics and proteomics data, respectively.
In the multi-omics fine-tuning stage, we fine-tuned the pre-trained model by feeding it thirty percent labeled data. After reaching convergence, the model generated global cell embeddings for the CITE-seq dataset and local cell embeddings for each respective data modality including transcriptomics and proteomics, which were prepared for downstream tasks. Finally, we used transcriptomics data to validate our model and followed a training process similar to the second stage.

\input{rnaseq_tab}

\subsubsection{Baseline Methods}
To demonstrate the usefulness of scFusionTTT, we applied several benchmark methods—CiteFuse, totalVI, SCOIT, jointDIMMSC, scMM, and BREMSC—to embed CITE-seq data in a common latent space during the multi-omics fine-tuning stage. Additionally, we used Scanpy, Seurat, Pagoda2, and scVI for single-modal prediction tasks. Due to the incomplete and unusable code for the scCTCLust method, it was not included in the comparison.
\begin{itemize}
\item CiteFuse \cite{kim2020citefuse} is a method that includes tools for pre-processing, modality integration, clustering, and RNA, protein expression analysis.

\item BREMSC \cite{wang2020brem} using Bayesian regression, which offers a robust and scalable method for modeling enhancer activity in transcriptomics.

\item jointDIMMSC \cite{sun2018dimm} is a method of joint dimensionality reduction that fuses transcriptomics and proteomics.

\item scMM \cite{minoura2021mixture} is based on a probabilistic mixture model framework and offers a robust method for analyzing and interpreting CITE-seq data.

\item SCOIT \cite{wang2023probabilistic}, a probabilistic tensor decomposition method, specifically designed to learn the unified embeddings from paired single-cell multi-omic data.

\item Total Variational Inference (TotalVI) \cite{gayoso2021joint} is a flexible generative model of CITE-seq dataset that can subsequently be used for many common downstream tasks.

\item Scanpy \cite{virshup2023scverse, li2024analysis} is a highly efficient and scalable toolkit for the analysis and visualization of transcriptomics data. It is designed to handle large-scale datasets and provides a comprehensive suite of tools for various stages of the scRNA-seq workflow, from preprocessing to downstream analysis. 

\item Seurat \cite{hao2024dictionary} is a comprehensive and widely-used toolkit for transcriptomics data analysis and visualization. It is designed to handle various stages of the scRNA-seq workflow, from data preprocessing to downstream analysis, and is particularly known for its versatility and robustness.

\item Pagoda2 \cite{barkas2021pagoda2}, a sophisticated computational framework designed for the analysis and visualization of transcriptomics data, which can handle large datasets and provide detailed insights into cellular heterogeneity and gene expression patterns.

\item Single-cell Variational Inference (scVI) \cite{gayoso2022python} is designed for the analysis of transcriptomics data. This method leverages a VAE model to effectively handle and analyze high-dimensional and sparse single-cell gene expression data through unsupervised learning. 

\end{itemize}
\subsubsection{Implementation Details}
In our research, we used a global cell embedding matrix $K \times N $ to represent low-dimensional embeddings of $K$ cells, the cell embedding comes from the results of scFusionTTT. These embeddings were then used for future investigations, which included creating a cell adjacency matrix and performing cell clustering. The adjacency matrix was generated from the cell embeddings using the K-nearest neighbors technique, with the number of neighbors set to the default value of 20. Cell clustering was subsequently performed using the Leiden method \cite{traag2019louvain}, which works with the adjacency matrix. To facilitate visualization, we used the Uniform Manifold Approximation and Projection (UMAP) technique to compress the cell embeddings and retrieve latent features to a two-dimensional space, allowing for visual identification of gene and protein expression levels across cell clusters.

\subsubsection{Clustering Performance}
To evaluate the performance of scFusionTTT, we employed 11 clustering metrics including the Adjusted Rand Index (ARI), the Normalised Mutual Information (NMI), Fowlkes-Mallows index (FMI), average silhouette width (ASW), adjusted mutual information (AMI), Jaccard index (JI), silhouette coefficient (SC), Calinski-Harabaz index (CHI), F-measure, and Davies-Bouldin index (DBI) to evaluate clustering results, which were internal and external indicators of clustering, such as the relationship of points within a class, the relationship between classes, the concentration of points, the distribution of points, and so on to give a very comprehensive assessment of clustering. Except DBI metric, the higher the value of the metric, the better the clustering performance.  Of note, some clustering metrics were not applicable to methods like BREMSC and jointDIMMSC, because these methods relied on coordinate-based visualization for their computations and do not generate independent visualizations.

\begin{figure*}[t]
  \centering
  \includegraphics[width=\textwidth]{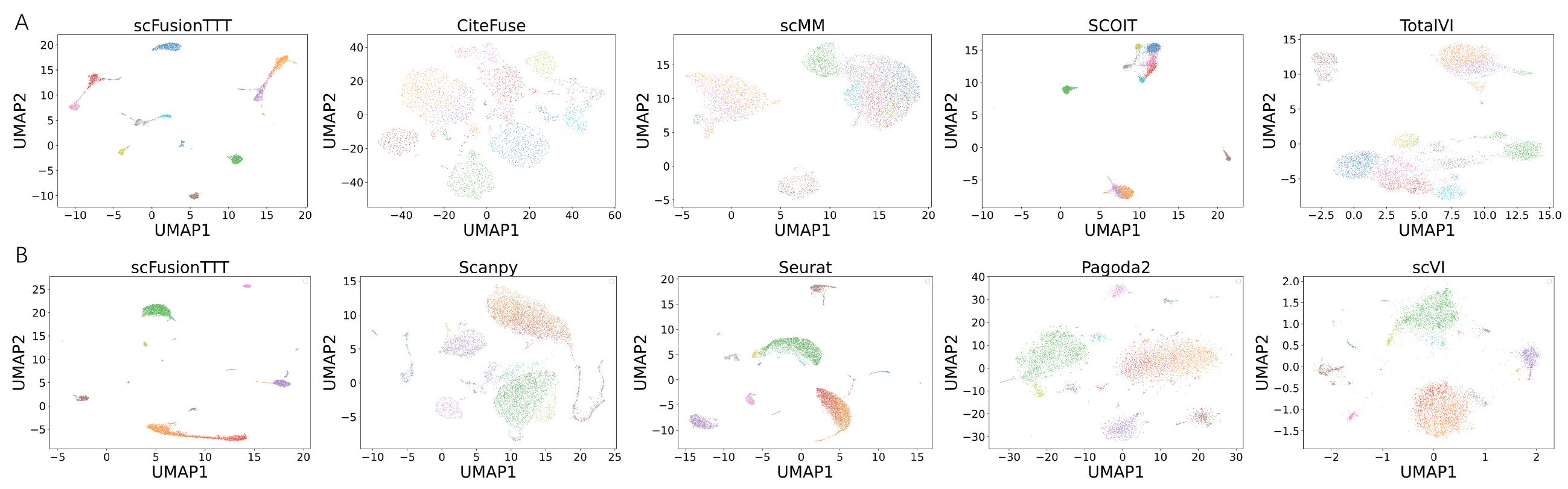}
  \vspace{-2em}
  \caption{(A) Comparison of UMAP visualization on PBMC10K CITE-seq dataset across different methods. (B) Comparison of UMAP visualization on CBMC transcriptomics dataset across different methods.}
  \vspace{-1em}
  \label{umap}
\end{figure*}

The clustering performance of our method compared to
the baseline methods on 4 CITE-seq datasets are presented in Table \ref{citeseq_tab}. Our model suFusionTTT got the state-of-the-art (SOTA) performance compared to other benchmarks in ten metrics across all CITE-seq datasets except for the JI metric in two datasets. Next, we used fine-tuned scFusionTTT to predict cell types of transcriptomics in stage 3. The clustering performance of our method compared to other methods on 4 scRNA-seq datasets is presented in Table \ref{rnaseq_tab}. Except for achieving the second-best results in the NMI and AMI metrics on the PBMC3K dataset and the ASW metric on the BMCITE dataset, scFusionTTT achieved the top performance in 10 other metrics across all experimental transcriptomics datasets.

Finally, we applied UMAP to visualize the cell embedding of scFusionTTT and the other baseline methods in CITE-seq datasets and scRNA-seq datasets with ground truth labels, as depicted in Figure \ref{umap}.  We observed that different classes of cells in the dataset can be well separated and matched with the ground truth well in the latent embedding representation generated by scFusionTTT. In summary, scFusionTTT could perform better than other methods, and the other datasets of UMAP visualization could be seen in Supplementary Figures.

\begin{figure}[t]
  \centering
  \vspace{-1em}
  \includegraphics[width=\columnwidth]{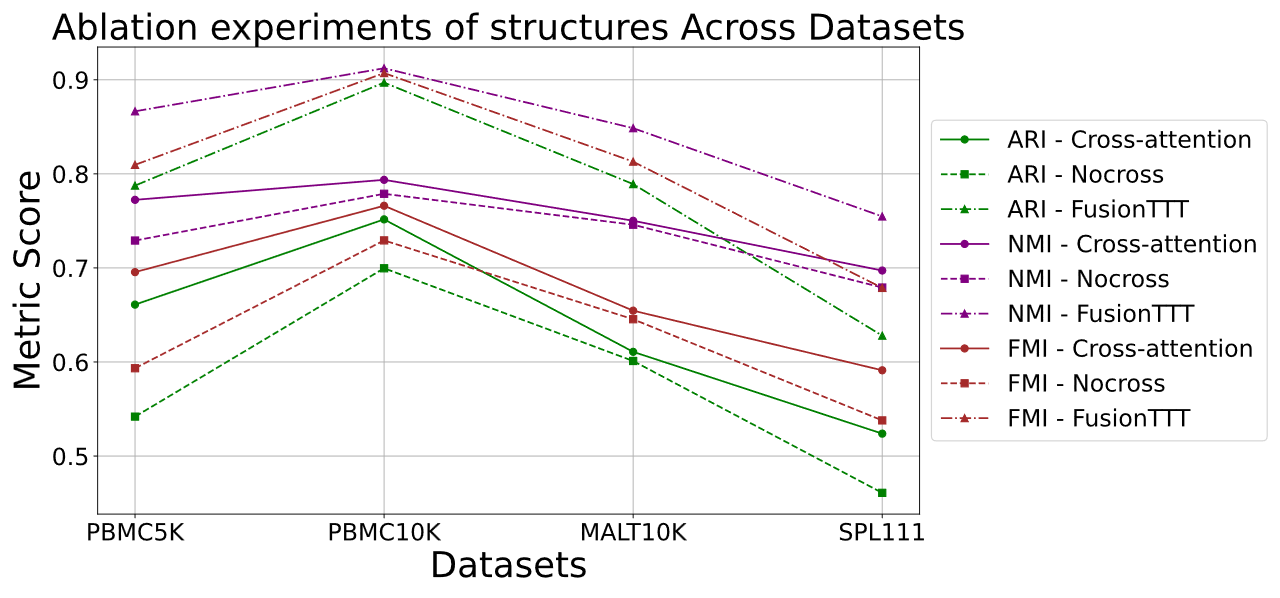}
  \vspace{-2em}
  \caption{ Performance of different fusion modules and without fusion mechanism evaluated by ARI, NMI, and FMI. ARI: adjusted rand index, NMI: normalized mutual information, FMI: Fowlkes-Mallows index.}
  \vspace{-1.5em}
  \label{ablation}
\end{figure}

\subsubsection{Ablation Study} 
In this experiment, we analyzed the impact of the TTT-based module. We specifically ablated the FusionTTT module and used the attention-based structure and corresponding element addition mechanism to compare it as shown in Figure \ref{ablation} evaluated by three main clustering evaluation metrics, whose results demonstrated better performance of our proposed module FusionTTT.

% In addition, We used the most mainstream attention mechanisms as components of the encoder and decoder to compare them with the TTT-based encoder and decoder. 
The TTT-based model had a better improvement compared to the attention-based structure and had a significant increase compared to the element addition mechanism. 
%We also conducted a gird-search to find the best parameters and suitable hyper-parameters including weight parameters. In summary, each aspect of the scFusionTTT is reasonable and valid.

% In addition, we conducted ablation studies on five CITE-seq datasets to underscore the effectiveness of the cross-attention architecture. We utilized three distinct evaluation metrics (ARI, NMI, and FMI) to determine the impact of cross-attention-based compared to the baseline model, which relies solely on element-wise addition (Fig ). The results showed that the indicators increased substantially on all datasets except NMI on the SPL206 dataset. These findings provide further evidence of the effectiveness of the proposed structural innovation. Additionally, we conducted mask ratio ablation experiments on five CITE-seq datasets, evaluated using ARI, NMI, and FMI, revealing that a 15 percent mask ratio yields optimal results (Fig ).

\subsubsection{Scalability of ScFusionTTT}
Due to the specificity of the scFusionTTT,  we could easily expand the modalities to three or more modalities, and choose any one of them as the main modality according to scenario requirements. %加一些三模态数据集的结果，如果来得急的话，

\subsubsection{Hyper-parameter Analysis}   
In this part, we determined the hyper-parameters $\alpha$ and $\beta$, optimal distribution of loss function in our model by gird-search. %The experiment process as shown in Supplementary Figure 1. %prioritizing transcriptomic data (0.7) over proteomic data (0.3).
In addition, the dimension $d$ of final cell representation is set to 128. % 不同数据集最优参数不一样，应该只选最优的 不给出最优数值？

%\subsubsection{Visualization Analysis}   To further demonstrate the superiority of scFusionTTT intuitively, we conduct Uniform Manifold Approximation and Projection (UMAP) on the cell embedding of all methods including scFusionTTT. It is observed that our scFusionTTT can better reveal the cell cluster compared with other baselines.

\subsubsection{Gene and Protein Order Analysis}   
\begin{figure}[t]
  \centering
  \vspace{-1em}
  \includegraphics[width=\columnwidth]{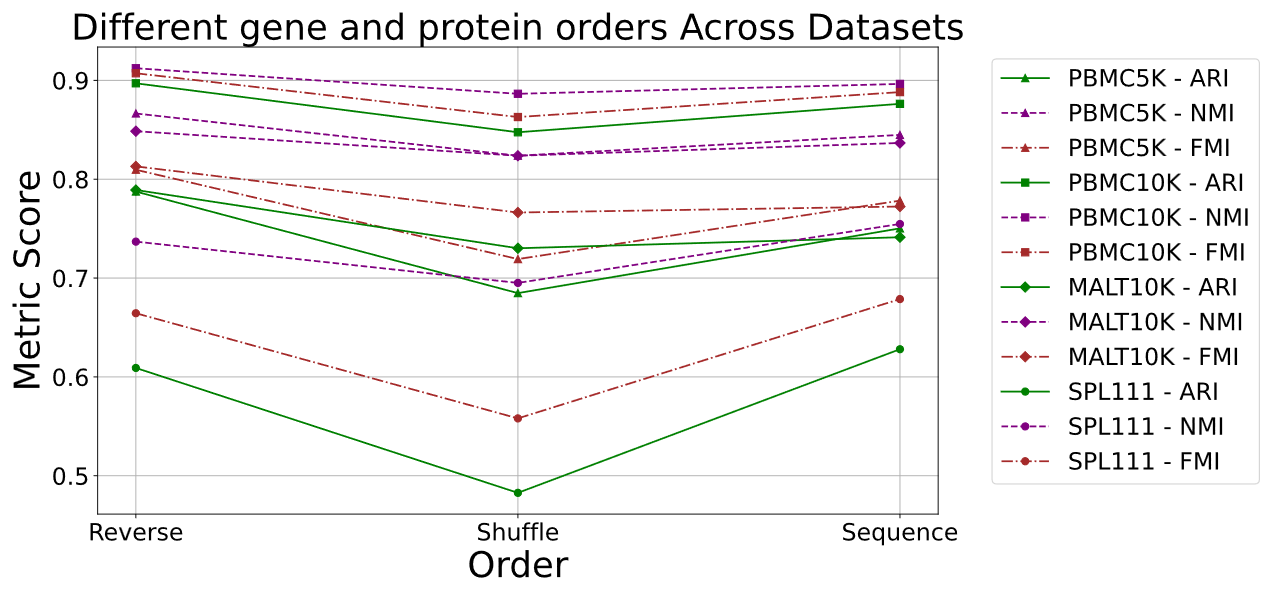}
  \vspace{-2em}
  \caption{Performance of different gene and protein orders evaluated by ARI, NMI, and FMI. ARI: adjusted rand index, NMI: normalized mutual information, FMI: Fowlkes-Mallows index}
  \vspace{-1.5em}
  \label{gene}
\end{figure}

In addition, we also analyzed the effect of gene and protein order on clustering results. We set up three scenarios, gene and protein in order, gene and protein in reverse order, and gene and protein in disrupted order as shown in Figure \ref{gene}. We observed that there is a significant decrease in the clustering results when we shuffled the gene and protein order, which proved our conjecture that the order information of genes and proteins in the genome is helpful for the final representations.

\section{Conclusion} 
In this paper, we propose a single-cell multi-omics fusion model scFusionTTT for clustering. The core idea is to utilize gene and protein order information in the human genome, followed by a TTT-layer updating mechanism that fits the sequence with a cause-and-effect relationship well. The respective representations are obtained from the output of the ADT encoder and RNA encoder. Then, each of the two FusionTTT modules fuses information unique to its respective modality and retains the information of the current modality, to reconstruct the expression matrix. Finally, the outputs of fusion modules will be regarded as the final representations to conduct downstream tasks including clustering.

Through experiments on 4 real CITE-seq datasets and 4 real transcriptomics datasets, we demonstrate the superiority of the proposed scFusionTTT method over other state-of-the-art baseline methods across most metrics in 8 datasets. Moreover, evidence from ablation studies and scalability studies demonstrates that scFusionTTT is robust, reliable, and extensible. The weakness of scFusionTTT is that it is not a cross-modal fusion, resulting in a weaker explanation of the relationships between the genes and proteins. Therefore, we will try to use the final representation to do some downstream tasks that reveal biological processes with better interpretation in the near future.

\bibliography{aaai25}

%%%%%%%%%%%%%%%%%%%%%%%%%%%%%%%%%%%%%%%%%%%%%%%%%%%%%%%%%%%%

\clearpage

\end{document}

\title{Supplementary Materials of scFusionTTT}
\maketitle
\vspace{-1.5em}
\section{Data availability}
The SPL111 CITE-seq dataset can be downloaded from \url{https://www.nature.com/articles/s41592-020-01050-x}. The PBMC5K, PBMC10K, MALT10K CITE-seq datasets are available at \url{https://support.10xgenomics.com/single-cell-gene-expression/datasets/3.0.2/5k_pbmc_protein_v3}, \url{https://support.10xgenomics.com/single-cell-gene-expression/datasets/3.0.0/pbmc_10k_v3}, and \url{https://support.10xgenomics.com/single-cell-gene-expression/datasets/3.0.0/malt_10k_protein_v3}, respectively.
The CBMC dataset is available at \url{https://www.nature.com/articles/nmeth.4380}, the PBMC3K dataset can be downloaded from \url{https://support.10xgenomics.com/single-cell-gene-expression/datasets/1.1.0/pbmc3k}, the BMCITE dataset sources from \url{https://www.cell.com/cell/fulltext/S0092-8674(21)00583-3}, and IFNB dataset can be downloaded from \url{https://www.nature.com/articles/nbt.4042}.
\vspace{-1.5em}
\section{Supplementary Methods}
\subsection{Sorting Algorithm}
We first used the dataset from the UCSC Genome Browser to generate a gene order list for the whole human genome. Then, we designed a sorting algorithm to align the experimental genes according to this genome-wide gene order list. Due to the diversity of gene aliases, we put all genes that do not map to the entire gene list table at the top. As for proteins, we sorted them in the order of the genes that are primarily translated into that protein.
 \vspace{-1.em}
\subsection{Random Masking}
The first stage of our model involves pre-training using self-supervised learning to capture gene and protein expression information. Based on experiments with different masking ratios, we selected a 15 percent masking ratio for our experiment.  In stage 2 and stage 3, we just use decoders to decode gene and protein data without masking. Of note, our masking behavior is after sorting genes and proteins. We randomly mask a portion of the gene and protein expression information, allowing the encoder to process only the unmasked portion. In the decoding stage, we combine the masked information from the initial step with the encoder's output and then feed this combined data into the decoder.
\section{Supplementary Figures}
\begin{figure}[H]
  \centering
  \includegraphics[width=\linewidth]{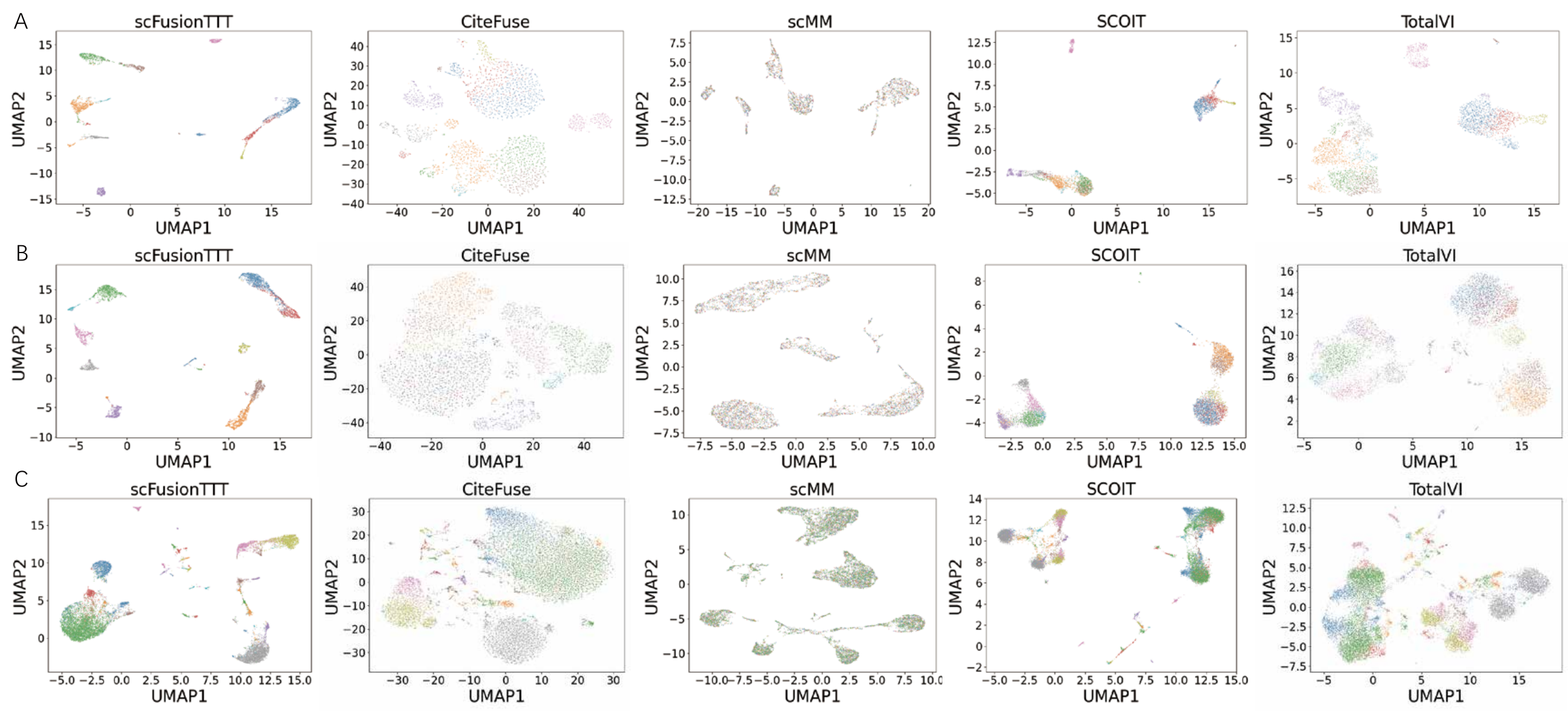}
  \caption{The UMAP visualization of other three CITE-seq datasets. (A) The UMAP visualization of PBMC5K dataset was generated by suFusionTTT, CiteFuse, scMM, SCOIT, and TotalVI, respectively. (B) The UMAP visualization of MALT10K dataset was generated by suFusionTTT, CiteFuse, scMM, SCOIT, and TotalVI, respectively. (C) The UMAP visualization of SPL111 dataset was generated by suFusionTTT, CiteFuse, scMM, SCOIT, and TotalVI, respectively.}
  \vspace{-2em}
\end{figure}

\begin{figure}[H]
  \centering
  \includegraphics[width=\linewidth]{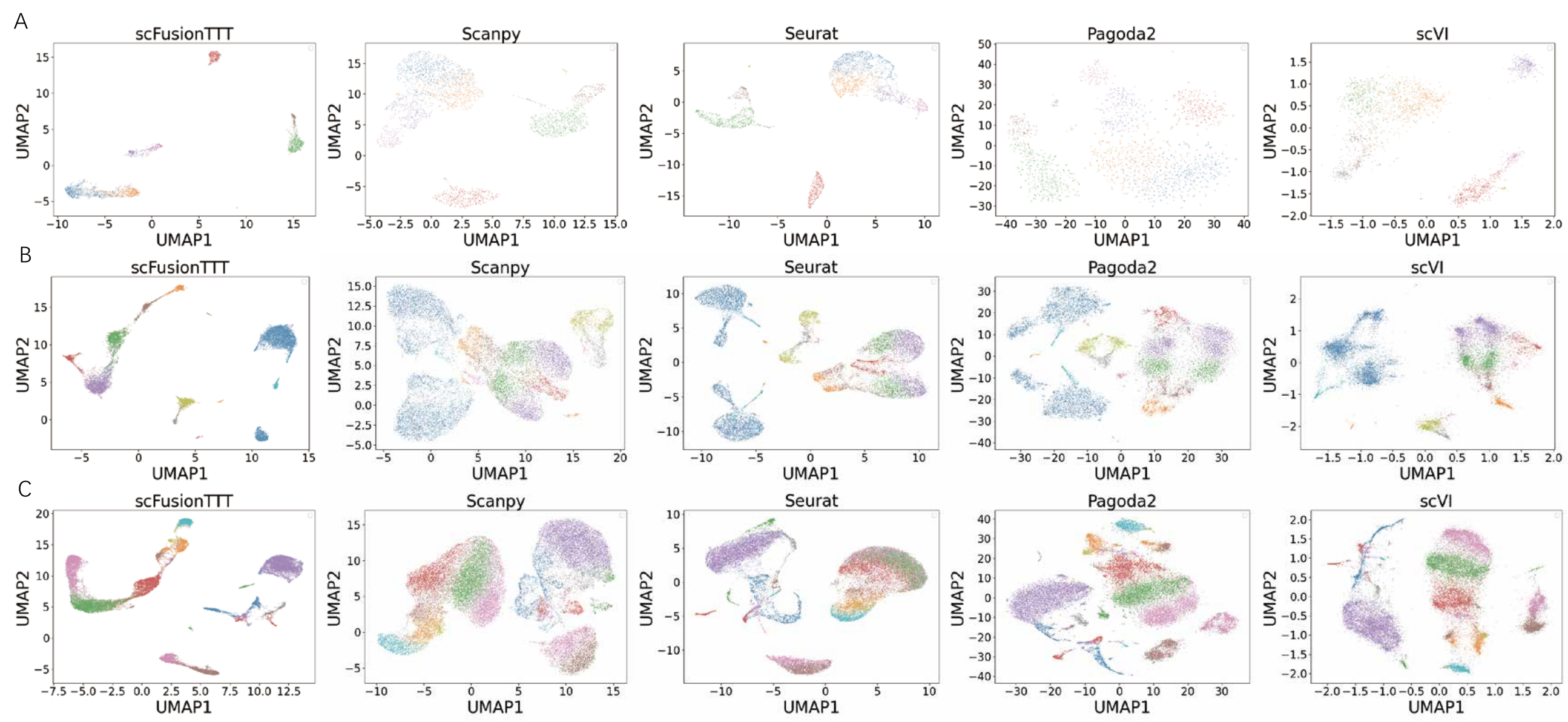}
  % \vspace{-2em}
  \caption{The UMAP visualization of other three scRNA-seq datasets. (A) The UMAP visualization of PBMC3K dataset generated by suFusionTTT, Scanpy, Seurat, Pagoda2, and scVI, respectively. (B) The UMAP visualization of IFNB dataset generated by suFusionTTT, Scanpy, Seurat, Pagoda2, and scVI, respectively. (C) The UMAP visualization of BMCITE dataset generated by suFusionTTT, Scanpy, Seurat, Pagoda2, and scVI, respectively.}
  \vspace{-1.5em}
  
\end{figure}

\bibliography{aaai25}

%% file: citeseq_tab.tex
\begin{table*}[t]

    \small
  \centering

  % \hspace{-2cm}
  \begin{adjustbox}{max width=\textwidth}
  \begin{tabular}{cccccccccccc}
    \toprule
    Dataset & Method & ARI $\uparrow$ & NMI $\uparrow$& FMI $\uparrow$& ASW $\uparrow$& AMI $\uparrow$& JI $\uparrow$& SC $\uparrow$& CHI $\uparrow$& F-measure $\uparrow$& DBI $\downarrow$\\
    \midrule
    \multirow{6}{*}{SPL111} & \cellcolor{lightgray}\textbf{scFusionTTT} & \cellcolor{lightgray}\textbf{0.63} & \cellcolor{lightgray}\textbf{0.75} & \cellcolor{lightgray}\textbf{0.68} & \cellcolor{lightgray}\textbf{0.66} & \cellcolor{lightgray}\textbf{0.75} & \cellcolor{lightgray}\textbf{0.51} & \cellcolor{lightgray}\textbf{0.50} & \cellcolor{lightgray}\textbf{36458} & \cellcolor{lightgray}\textbf{0.67} & \cellcolor{lightgray}\textbf{1.10} \\

                            & CiteFuse & 0.55 &  0.64 & 0.63 & 0.49 & 0.64 & 0.45& 0.30 & 9504 & 0.62 & 1.20 \\
                            & BREMSC & 0.31 & 0.58 & 0.39 & - & 0.58 & 0.23 & - & - & 0.37 & - \\
                            & JointDIMMSC & 0.32 & 0.59 & 0.40 & - & 0.58 & 0.24 & - & - & 0.39 & - \\
                            & scMM & -0.01 & 0.01 & 0.10 & 0.41 & 0.51 & 0.05 & 0.42 & 33011 & 0.09 & 1.17 \\
                            & SCOIT & 0.37 & 0.59 & 0.45 & 0.43 & 0.58 & 0.29 & 0.06 & 493 & 0.45 & 2.81 \\
                            & TotalVI & 0.39 & 0.64 & 0.47 & 0.44 & 0.64 & 0.30 & 0.39 & 20615 & 0.46 & 1.35 \\

    \hline
    \multirow{7}{*}{PBMC5K} & \cellcolor{lightgray}\textbf{scFusionTTT} & \cellcolor{lightgray}\textbf{0.79} & \cellcolor{lightgray}\textbf{0.87} & \cellcolor{lightgray}\textbf{0.81} & \cellcolor{lightgray}\textbf{0.76} & \cellcolor{lightgray}\textbf{0.86} & \cellcolor{lightgray}\textbf{0.67} & \cellcolor{lightgray}\textbf{0.68} & \cellcolor{lightgray}\textbf{48752} & \cellcolor{lightgray}\textbf{0.81} & \cellcolor{lightgray}\textbf{0.39} \\
                            & CiteFuse & 0.65 & 0.78 & 0.68 & 0.64 & 0.78 & 0.51& 0.42 & 3840 & 0.68 & 0.77 \\
                            & BREMSC & 0.54 & 0.72 & 0.61 & - & 0.75 & 0.43 & - & - & 0.60 & - \\
                            & JointDIMMSC & 0.48 & 0.71 & 0.59 & - & 0.71 & 0.38 & - & - & 0.55 & - \\
                            & scMM & 0.01 & 0.01 & 0.08 & 0.46 & 0.01 & 0.04 & 0.48 & 35795 & 0.09 & 0.65 \\
                            & SCOIT & 0.55 & 0.75 & 0.65 & 0.62 & 0.75 & {0.45} & 0.12 & 235 & 0.62 & 2.17 \\
                            & TotalVI & 0.65 & 0.80 & 0.69 & 0.64 & 0.80 & 0.52 & 0.38 & 11529 & 0.68 & 0.74 \\
                                                        
    \hline
    \multirow{7}{*}{PBMC10K} & \cellcolor{lightgray}\textbf{scFusionTTT} & \cellcolor{lightgray}\textbf{0.90} & \cellcolor{lightgray}\textbf{0.91} & \cellcolor{lightgray}\textbf{0.91} & \cellcolor{lightgray}\textbf{0.81} & \cellcolor{lightgray}\textbf{0.91} & \cellcolor{lightgray}{0.83} & \cellcolor{lightgray}\textbf{0.72} & \cellcolor{lightgray}\textbf{43915} & \cellcolor{lightgray}\textbf{0.91} & \cellcolor{lightgray}\textbf{0.48} \\
                             & CiteFuse & 0.79 &  0.84 & 0.81 & 0.64 & 0.84 & 0.68& 0.34 & 5945 & \textbf{0.91} & 1.00 \\
                             & BREMSC & 0.70 & 0.79 & 0.74 & - & 0.79 & 0.58 & - & - & 0.73 & - \\
                        
                             & JointDIMMSC & 0.54 & 0.74 & 0.59 & - & 0.74 & 0.41 & - & - & 0.59 & - \\
                             & scMM & 0.00 & 0.01 & 0.09 & 0.48 & 0.00 & 0.04 & 0.47 & 26481 & 0.09 & {0.76} \\
                             & SCOIT & 0.68 & 0.79 & 0.73 & 0.59 & 0.79 & \textbf{1.00} & 0.11 & 354 & 0.72 & 2.41 \\
                             & TotalVI & 0.71 & {0.81} & 0.74 & 0.63 & 0.81 & 0.58 & 0.36 & 15955 & 0.74 & 1.60 \\
                              
    \hline
    \multirow{7}{*}{MALT10K} & \cellcolor{lightgray}\textbf{scFusionTTT} & \cellcolor{lightgray}\textbf{0.79} & \cellcolor{lightgray}\textbf{0.85} & \cellcolor{lightgray}\textbf{0.81} & \cellcolor{lightgray}\textbf{0.78} & \cellcolor{lightgray}\textbf{0.85} & \cellcolor{lightgray}{0.68} & \cellcolor{lightgray}\textbf{0.68} & \cellcolor{lightgray}\textbf{62325} & \cellcolor{lightgray}\textbf{0.81} & \cellcolor{lightgray}\textbf{0.38} \\
                             & CiteFuse & 0.58 &  0.74 & {0.64} & {0.57} & 0.74 & {0.46}& 0.21 & 3705 & 0.63 & 1.10 \\
                             & BREMSC & 0.52 & 0.70 & 0.57 & - & 0.70 & 0.40 & - & - & 0.57 & - \\
                             & JointDIMMSC & 0.55 & 0.71 & 0.63 & - & 0.71 & 0.44 & - & - &{0.61} & - \\
                             & scMM & -0.00 & 0.01 & 0.10 & 0.48 & -0.00 & 0.05 & 0.32 & {18645} & 0.10 & {0.96} \\
                             & SCOIT & 0.53 & 0.71 & {0.64} & 0.55 & 0.71 & \textbf{1.00} & 0.08 & 266 & 0.60 & 2.85 \\
                             & TotalVI & {0.57} & {0.73} & 0.62 & {0.59} & {0.73} & 0.45 & {0.34} & 14229 & 0.62 & 0.87 \\
                             
    \bottomrule
  \end{tabular}
  \end{adjustbox}
  \vspace{-1em}
    \caption{Ten evaluation metric scores for the five methods on the four CITE-seq datasets. The best results in each column are bolded. - represents that this metric is not applicable to this method.}
    \vspace{-2em}
  \label{citeseq_tab}
\end{table*}

%% file: rnaseq_tab.tex
\begin{table*}[t]

    % \vspace{-0.5em}
    \small
  \centering

  % \hspace{-1cm}
  \begin{adjustbox}{max width=\textwidth}

  \begin{tabular}{cccccccccccc}
    
    \toprule
    Dataset & Method & ARI $\uparrow$& NMI $\uparrow$& FMI $\uparrow$& AMI $\uparrow$& ASW $\uparrow$& CHI $\uparrow$& F-measure $\uparrow$& JI $\uparrow$& SC $\uparrow$& DBI $\downarrow$\\
    \midrule
    \multirow{5}{*}{IFNB} & \cellcolor{lightgray}\textbf{scFusionTTT} & \cellcolor{lightgray}\textbf{0.61} & \cellcolor{lightgray}\textbf{0.77} & \cellcolor{lightgray}\textbf{0.68} & \cellcolor{lightgray}\textbf{0.77} & \cellcolor{lightgray}\textbf{0.79} & \cellcolor{lightgray}\textbf{55698} & \cellcolor{lightgray}\textbf{0.66} & \cellcolor{lightgray}\textbf{0.49} & \cellcolor{lightgray}\textbf{0.58} & \cellcolor{lightgray}\textbf{0.77} \\
                          & Scanpy & 0.39 & 0.67 & 0.50 & 0.67 & 0.50 & 18222 & 0.45 & 0.29 & 0.37 & 1.06 \\
                          & Seurat & 0.46 & 0.67 & 0.56 & 0.67 & 0.51 & 31037 & 0.52 & 0.35 & 0.36 & 1.31 \\
                          & Pagoda2 & 0.25 & 0.63 & 0.39 & 0.63 & 0.53 & 8069 & 0.29 & 0.17 & 0.29 & 2.89 \\
                          & scVI & 0.54 & 0.72 & 0.61 & 0.72 & 0.68 & 17278 & 0.60 & 0.43 & 0.43 & 1.21 \\
    \hline
    \multirow{5}{*}{PBMC3K} & \cellcolor{lightgray}\textbf{scFusionTTT} & \cellcolor{lightgray}\textbf{0.66} & \cellcolor{lightgray}{0.72} & \cellcolor{lightgray}\textbf{0.71} & \cellcolor{lightgray}{0.71} & \cellcolor{lightgray}\textbf{0.72} & \cellcolor{lightgray}\textbf{16302} & \cellcolor{lightgray}\textbf{0.71} & \cellcolor{lightgray}\textbf{0.55} & \cellcolor{lightgray}\textbf{0.63} & \cellcolor{lightgray}\textbf{0.49} \\
                            & Scanpy & 0.53 & 0.71 & 0.61 & 0.70 & 0.70 & 7458 & 0.59 & 0.42 & 0.38 & 0.81 \\
                            & Seurat & 0.61 & \textbf{0.77} & 0.68 & \textbf{0.77} & 0.71 & 10215 & 0.66 & 0.49 & 0.48 & 0.64 \\
                            & Pagoda2 & 0.45 & 0.63 & 0.55 & 0.63 & 0.66 & 790 & 0.50 & 0.34 & 0.18 & 4.35 \\
                            & scVI & 0.42 & 0.62 & 0.51 & 0.61 & 0.66 & 3831 & 0.49 & 0.33 & 0.34 & 1.02 \\
    \hline
    \multirow{5}{*}{CBMC} & \cellcolor{lightgray}\textbf{scFusionTTT} & \cellcolor{lightgray}\textbf{0.76} & \cellcolor{lightgray}\textbf{0.80} & \cellcolor{lightgray}\textbf{0.80} & \cellcolor{lightgray}\textbf{0.80} & \cellcolor{lightgray}\textbf{0.77} & \cellcolor{lightgray}\textbf{60770} & \cellcolor{lightgray}\textbf{0.80} & \cellcolor{lightgray}\textbf{0.67} & \cellcolor{lightgray}\textbf{0.68} & \cellcolor{lightgray}\textbf{0.49} \\
                          & Scanpy & 0.34 & 0.64 & 0.44 & 0.64 & 0.56 & 7916 & 0.40 & 0.25 & 0.31 & 1.35 \\
                          & Seurat & 0.47 & 0.73 & 0.55 & 0.73 & 0.61 & 14628 & 0.54 & 0.37 & 0.44 & 1.39 \\
                          & Pagoda2 & 0.36 & 0.65 & 0.46 & 0.64 & 0.63 & 2978 & 0.42 & 0.27 & 0.20 & 8.96 \\
                          & scVI & 0.53 & 0.73 & 0.60 & 0.73 & {0.73} & 11215 & {0.59} &{0.41} & 0.38 & 0.73 \\
    \hline
    \multirow{5}{*}{BMCITE} & \cellcolor{lightgray}\textbf{scFusionTTT} & \cellcolor{lightgray}\textbf{0.71} & \cellcolor{lightgray}\textbf{0.77} & \cellcolor{lightgray}\textbf{0.74} & \cellcolor{lightgray}\textbf{0.77} & \cellcolor{lightgray}{0.71} & \cellcolor{lightgray}\textbf{105530} & \cellcolor{lightgray}\textbf{0.74} & \cellcolor{lightgray}\textbf{0.58} & \cellcolor{lightgray}\textbf{0.51} & \cellcolor{lightgray}\textbf{0.73} \\
                            & Scanpy & 0.46 & 0.68 & {0.52} & 0.68 & 0.56 & 34711 & {0.51} & {0.34} & 0.31 & 0.93 \\
                            & Seurat & 0.44 & 0.67 & 0.49 & 0.67 & 0.55 & {52840} & 0.49 & 0.33 & 0.36 & 0.85 \\
                            & Pagoda2 & 0.41 & {0.70} & 0.51 & {0.70} & {0.63} & 10929 & 0.45 & 0.29 & 0.26 & 3.13 \\
                            & scVI & {0.62} & \textbf{0.77} & {0.66} & \textbf{0.77} &\textbf{0.77} & 36178 & {0.65} & {0.48} & {0.43} & {0.84} \\
    \bottomrule
  \end{tabular}
  \end{adjustbox}
  \vspace{-1em}
  \caption{Ten evaluation metric scores for the five methods on the four transcriptomics datasets. The best results in each column are bolded.}
  \vspace{-2em}
  \label{rnaseq_tab}
\end{table*}

%% file: anonymous-submission-latex-2025.bbl
\begin{thebibliography}{32}
\providecommand{\natexlab}[1]{#1}

\bibitem[{Aitchison(1982)}]{aitchison1982statistical}
Aitchison, J. 1982.
\newblock The statistical analysis of compositional data.
\newblock \emph{Journal of the Royal Statistical Society: Series B (Methodological)}, 44(2): 139--160.

\bibitem[{Barkas et~al.(2021)Barkas, Petukhov, Kharchenko, and Biederstedt}]{barkas2021pagoda2}
Barkas, N.; Petukhov, V.; Kharchenko, P.; and Biederstedt, E. 2021.
\newblock pagoda2: single cell analysis and differential expression.
\newblock \emph{R package version}, 1(8).

\bibitem[{Boehm et~al.(2022)Boehm, Khosravi, Vanguri, Gao, and Shah}]{boehm2022harnessing}
Boehm, K.~M.; Khosravi, P.; Vanguri, R.; Gao, J.; and Shah, S.~P. 2022.
\newblock Harnessing multimodal data integration to advance precision oncology.
\newblock \emph{Nature Reviews Cancer}, 22(2): 114--126.

\bibitem[{Chen et~al.(2020)Chen, Fan, Girshick, and He}]{chen2020improved}
Chen, X.; Fan, H.; Girshick, R.; and He, K. 2020.
\newblock Improved baselines with momentum contrastive learning.
\newblock \emph{arXiv preprint arXiv:2003.04297}.

\bibitem[{Conesa et~al.(2016)Conesa, Madrigal, Tarazona, Gomez-Cabrero, Cervera, McPherson, Szcze{\'s}niak, Gaffney, Elo, Zhang et~al.}]{conesa2016survey}
Conesa, A.; Madrigal, P.; Tarazona, S.; Gomez-Cabrero, D.; Cervera, A.; McPherson, A.; Szcze{\'s}niak, M.~W.; Gaffney, D.~J.; Elo, L.~L.; Zhang, X.; et~al. 2016.
\newblock A survey of best practices for RNA-seq data analysis.
\newblock \emph{Genome biology}, 17: 1--19.

\bibitem[{Dandekar et~al.(1998)Dandekar, Snel, Huynen, and Bork}]{dandekar1998conservation}
Dandekar, T.; Snel, B.; Huynen, M.; and Bork, P. 1998.
\newblock Conservation of gene order: a fingerprint of proteins that physically interact.
\newblock \emph{Trends in biochemical sciences}, 23(9): 324--328.

\bibitem[{Doersch(2016)}]{doersch2016tutorial}
Doersch, C. 2016.
\newblock Tutorial on variational autoencoders.
\newblock \emph{arXiv preprint arXiv:1606.05908}.

\bibitem[{Gayoso et~al.(2022)Gayoso, Lopez, Xing, Boyeau, Valiollah Pour~Amiri, Hong, Wu, Jayasuriya, Mehlman, Langevin et~al.}]{gayoso2022python}
Gayoso, A.; Lopez, R.; Xing, G.; Boyeau, P.; Valiollah Pour~Amiri, V.; Hong, J.; Wu, K.; Jayasuriya, M.; Mehlman, E.; Langevin, M.; et~al. 2022.
\newblock A Python library for probabilistic analysis of single-cell omics data.
\newblock \emph{Nature biotechnology}, 40(2): 163--166.

\bibitem[{Gayoso et~al.(2021)Gayoso, Steier, Lopez, Regier, Nazor, Streets, and Yosef}]{gayoso2021joint}
Gayoso, A.; Steier, Z.; Lopez, R.; Regier, J.; Nazor, K.~L.; Streets, A.; and Yosef, N. 2021.
\newblock Joint probabilistic modeling of single-cell multi-omic data with totalVI.
\newblock \emph{Nature methods}, 18(3): 272--282.

\bibitem[{Gu and Dao(2023)}]{gu2023mamba}
Gu, A.; and Dao, T. 2023.
\newblock Mamba: Linear-time sequence modeling with selective state spaces.
\newblock \emph{arXiv preprint arXiv:2312.00752}.

\bibitem[{Hao et~al.(2024)Hao, Stuart, Kowalski, Choudhary, Hoffman, Hartman, Srivastava, Molla, Madad, Fernandez-Granda et~al.}]{hao2024dictionary}
Hao, Y.; Stuart, T.; Kowalski, M.~H.; Choudhary, S.; Hoffman, P.; Hartman, A.; Srivastava, A.; Molla, G.; Madad, S.; Fernandez-Granda, C.; et~al. 2024.
\newblock Dictionary learning for integrative, multimodal and scalable single-cell analysis.
\newblock \emph{Nature biotechnology}, 42(2): 293--304.

\bibitem[{Hurst, P{\'a}l, and Lercher(2004)}]{hurst2004evolutionary}
Hurst, L.~D.; P{\'a}l, C.; and Lercher, M.~J. 2004.
\newblock The evolutionary dynamics of eukaryotic gene order.
\newblock \emph{Nature Reviews Genetics}, 5(4): 299--310.

\bibitem[{Kim et~al.(2020)Kim, Lin, Geddes, Yang, and Yang}]{kim2020citefuse}
Kim, H.~J.; Lin, Y.; Geddes, T.~A.; Yang, J. Y.~H.; and Yang, P. 2020.
\newblock CiteFuse enables multi-modal analysis of CITE-seq data.
\newblock \emph{Bioinformatics}, 36(14): 4137--4143.

\bibitem[{Li et~al.(2024)Li, Sun, Romanova, Wu, Fang, and Moroz}]{li2024analysis}
Li, Y.; Sun, C.; Romanova, D.~Y.; Wu, D.~O.; Fang, R.; and Moroz, L.~L. 2024.
\newblock Analysis and Visualization of Single-Cell Sequencing Data with Scanpy and MetaCell: A Tutorial.
\newblock \emph{Ctenophores: Methods and Protocols}, 383--445.

\bibitem[{Maston, Evans, and Green(2006)}]{maston2006transcriptional}
Maston, G.~A.; Evans, S.~K.; and Green, M.~R. 2006.
\newblock Transcriptional regulatory elements in the human genome.
\newblock \emph{Annu. Rev. Genomics Hum. Genet.}, 7(1): 29--59.

\bibitem[{Miao et~al.(2021)Miao, Humphreys, McMahon, and Kim}]{miao2021multi}
Miao, Z.; Humphreys, B.~D.; McMahon, A.~P.; and Kim, J. 2021.
\newblock Multi-omics integration in the age of million single-cell data.
\newblock \emph{Nature Reviews Nephrology}, 17(11): 710--724.

\bibitem[{Minoura et~al.(2021)Minoura, Abe, Nam, Nishikawa, and Shimamura}]{minoura2021mixture}
Minoura, K.; Abe, K.; Nam, H.; Nishikawa, H.; and Shimamura, T. 2021.
\newblock A mixture-of-experts deep generative model for integrated analysis of single-cell multiomics data.
\newblock \emph{Cell reports methods}, 1(5).

\bibitem[{Peng et~al.(2023)Peng, Alcaide, Anthony, Albalak, Arcadinho, Biderman, Cao, Cheng, Chung, Grella et~al.}]{peng2023rwkv}
Peng, B.; Alcaide, E.; Anthony, Q.; Albalak, A.; Arcadinho, S.; Biderman, S.; Cao, H.; Cheng, X.; Chung, M.; Grella, M.; et~al. 2023.
\newblock Rwkv: Reinventing rnns for the transformer era.
\newblock \emph{arXiv preprint arXiv:2305.13048}.

\bibitem[{Salehinejad et~al.(2017)Salehinejad, Sankar, Barfett, Colak, and Valaee}]{salehinejad2017recent}
Salehinejad, H.; Sankar, S.; Barfett, J.; Colak, E.; and Valaee, S. 2017.
\newblock Recent advances in recurrent neural networks.
\newblock \emph{arXiv preprint arXiv:1801.01078}.

\bibitem[{Sherstinsky(2020)}]{sherstinsky2020fundamentals}
Sherstinsky, A. 2020.
\newblock Fundamentals of recurrent neural network (RNN) and long short-term memory (LSTM) network.
\newblock \emph{Physica D: Nonlinear Phenomena}, 404: 132306.

\bibitem[{Steier, Maslan, and Streets(2022)}]{steier2022joint}
Steier, Z.; Maslan, A.; and Streets, A. 2022.
\newblock Joint Analysis of Transcriptome and Proteome Measurements in Single Cells with totalVI.
\newblock In \emph{Single Cell ‘Omics of Neuronal Cells}, 63--85. Springer.

\bibitem[{Stoeckius et~al.(2017)Stoeckius, Hafemeister, Stephenson, Houck-Loomis, Chattopadhyay, Swerdlow, Satija, and Smibert}]{stoeckius2017simultaneous}
Stoeckius, M.; Hafemeister, C.; Stephenson, W.; Houck-Loomis, B.; Chattopadhyay, P.~K.; Swerdlow, H.; Satija, R.; and Smibert, P. 2017.
\newblock Simultaneous epitope and transcriptome measurement in single cells.
\newblock \emph{Nature methods}, 14(9): 865--868.

\bibitem[{Sun et~al.(2024)Sun, Li, Dalal, Xu, Vikram, Zhang, Dubois, Chen, Wang, Koyejo et~al.}]{sun2024learning}
Sun, Y.; Li, X.; Dalal, K.; Xu, J.; Vikram, A.; Zhang, G.; Dubois, Y.; Chen, X.; Wang, X.; Koyejo, S.; et~al. 2024.
\newblock Learning to (learn at test time): Rnns with expressive hidden states.
\newblock \emph{arXiv preprint arXiv:2407.04620}.

\bibitem[{Sun et~al.(2018)Sun, Wang, Deng, Wang, Lafyatis, Ding, Hu, and Chen}]{sun2018dimm}
Sun, Z.; Wang, T.; Deng, K.; Wang, X.-F.; Lafyatis, R.; Ding, Y.; Hu, M.; and Chen, W. 2018.
\newblock DIMM-SC: a Dirichlet mixture model for clustering droplet-based single cell transcriptomic data.
\newblock \emph{Bioinformatics}, 34(1): 139--146.

\bibitem[{Traag, Waltman, and Van~Eck(2019)}]{traag2019louvain}
Traag, V.~A.; Waltman, L.; and Van~Eck, N.~J. 2019.
\newblock From Louvain to Leiden: guaranteeing well-connected communities.
\newblock \emph{Scientific reports}, 9(1): 1--12.

\bibitem[{Vaswani(2017)}]{vaswani2017attention}
Vaswani, A. 2017.
\newblock Attention is all you need.
\newblock \emph{arXiv preprint arXiv:1706.03762}.

\bibitem[{Virshup et~al.(2023)Virshup, Bredikhin, Heumos, Palla, Sturm, Gayoso, Kats, Koutrouli, Berger et~al.}]{virshup2023scverse}
Virshup, I.; Bredikhin, D.; Heumos, L.; Palla, G.; Sturm, G.; Gayoso, A.; Kats, I.; Koutrouli, M.; Berger, B.; et~al. 2023.
\newblock The scverse project provides a computational ecosystem for single-cell omics data analysis.
\newblock \emph{Nature biotechnology}, 41(5): 604--606.

\bibitem[{Wang, Wang, and Li(2023)}]{wang2023probabilistic}
Wang, R.~H.; Wang, J.; and Li, S.~C. 2023.
\newblock Probabilistic tensor decomposition extracts better latent embeddings from single-cell multiomic data.
\newblock \emph{Nucleic acids research}, 51(15): e81--e81.

\bibitem[{Wang et~al.(2020)Wang, Sun, Zhang, Xu, Xin, Huang, Duerr, Chen, Ding, and Chen}]{wang2020brem}
Wang, X.; Sun, Z.; Zhang, Y.; Xu, Z.; Xin, H.; Huang, H.; Duerr, R.~H.; Chen, K.; Ding, Y.; and Chen, W. 2020.
\newblock BREM-SC: a bayesian random effects mixture model for joint clustering single cell multi-omics data.
\newblock \emph{Nucleic acids research}, 48(11): 5814--5824.

\bibitem[{Wittkopp and Kalay(2012)}]{wittkopp2012cis}
Wittkopp, P.~J.; and Kalay, G. 2012.
\newblock Cis-regulatory elements: molecular mechanisms and evolutionary processes underlying divergence.
\newblock \emph{Nature Reviews Genetics}, 13(1): 59--69.

\bibitem[{Yuan, Chen, and Deng(2022)}]{yuan2022clustering}
Yuan, M.; Chen, L.; and Deng, M. 2022.
\newblock Clustering CITE-seq data with a canonical correlation-based deep learning method.
\newblock \emph{Frontiers in Genetics}, 13: 977968.

\bibitem[{Zhang and Sennrich(2019)}]{zhang2019root}
Zhang, B.; and Sennrich, R. 2019.
\newblock Root mean square layer normalization.
\newblock \emph{Advances in Neural Information Processing Systems}, 32.

\end{thebibliography}
